\documentclass[conference]{IEEEtran}
\IEEEoverridecommandlockouts
\usepackage{cite}
\usepackage{amsmath,amssymb,amsfonts}
\usepackage{algorithmic}
\usepackage{graphicx}
\usepackage{textcomp}
\usepackage{xcolor}
\def\BibTeX{{\rm B\kern-.05em{\sc i\kern-.025em b}\kern-.08em
    T\kern-.1667em\lower.7ex\hbox{E}\kern-.125emX}}
\begin{document}

\title{Towards Resource Efficient and Interpretable Bias Mitigation in Large Language Models}

\author{
    \IEEEauthorblockN{
    1\textsuperscript{st} Schrasing Tong\IEEEauthorrefmark{1},
    1\textsuperscript{st} Eliott Zemour\IEEEauthorrefmark{1}\IEEEauthorrefmark{2},
    3\textsuperscript{rd} Jessica Lu\IEEEauthorrefmark{1}, 4\textsuperscript{th} Rawisara Lohanimit\IEEEauthorrefmark{1}, 5\textsuperscript{th} Lalana Kagal\IEEEauthorrefmark{1}}
    \IEEEauthorblockA{\IEEEauthorrefmark{1}
        Department of Electrical Engineering and Computer Science \\
        Massachusetts Institute of Technology \\
        Cambridge, MA, USA \\
        \{st9, eliozem, jlu25, rloha\}@mit.edu, \{lkagal\}@csail.mit.edu
    }
    \IEEEauthorblockA{\IEEEauthorrefmark{2}
        Dynamo AI \\
        San Francisco, CA, USA \\
        \{eliott\}@dynamo.ai
    }
}


\maketitle

\begin{abstract}
Although large language models (LLMs) have demonstrated their effectiveness in a wide range of applications, they have also been observed to perpetuate unwanted biases present in the training data, potentially leading to harm for marginalized communities. 
In this paper, we mitigate bias by leveraging small biased and anti-biased expert models to obtain a debiasing signal that is added to the LLM output at decoding-time. 
This approach combines computational efficiency - fine-tuning a small model versus re-training a large model and interpretability - one can examine the probability shift from debiasing.  
The framework can also be tailored to specific contexts by switching the choice of the fine-tuning dataset. 
Experiments on mitigating gender, race, and religion biases on different architectures show a reduction in bias on several local and global bias metrics while preserving language model performance.
\end{abstract}


\section{Introduction}
Natural language generation (NLG) has become increasingly popular, serving as building blocks for applications such as chatbots, translators, and writing assistants and interacting with users in different domains~\cite{gatt2018survey}. 
However, despite recent advances, these large language models (LLMs) have been reported to capture and reproduce unwanted biases and stereotypes~\cite{abid2021persistent,lucy-bamman-2021-gender}.
This occurs mainly because the large text corpora required for training such models are extracted from the Web, which is not an accurate reflection of the diversity of real-world distributions.
The generation of biased output can have serious negative consequences for society, ranging from offensive language that prevents certain demographic groups from adopting technology~\cite{Sheng2020} to biased job ads that discourage candidates from applying to certain positions~\cite{borchers-etal-2022-looking}.

To address these issues, researchers have tried to curate better training data and improve the training process~\cite{Dinan2019,Liu2019cda}. 
Nevertheless, this approach remains less feasible in practice due to the significant human and computation resources involved. 
Recent works have instead focused on reducing the bias of generated outputs at decoding time to improve computational efficiency.
For example,~\cite{Sheng2020} introduced a prompt engineering method, Trigger, which concatenates a sequence of tokens to user inputs to reduce the output's bias. 
However, the modified prompts suffer from a lack of interpretability and the system has been shown to spew racist output on non-racial contexts~\cite{Wallace2021}.

In this paper, we adapt the detoxification methodology that uses small fine-tuned models~\cite{Liu2021} to mitigating bias.
We use small language models (LMs), for instance pre-trained GPT-2 Small or LLaMA 3.2 1B models, fine-tuned on subsets of the RedditBias dataset~\cite{Barikeri2021}, as the biased and anti-biased experts in our system. 
They produce a debiasing signal that is incorporated into the target LLM output at decoding-time; this LLM can have any architecture, provided it shares the same dictionary as the experts.  
This approach is computationally efficient because fine-tuning offers significant cost savings compared to re-training the entire model, and fine-tuning smaller expert models is more efficient than directly fine-tuning the target model. 
It is also interpretable, as one can examine the shift in output probabilities for any given prompt.
Furthermore, if the context and cause of bias are known beforehand, the framework can be easily adjusted by fine-tuning with a targeted dataset. 

We evaluated the performance of the framework on two different target models along three different bias directions - gender, race, and religion. 
We observed a reduction in bias on several local and global bias metrics such as Regard~\cite{Sheng2019}, Toxicity, Hellinger distance, and Stereotype Score~\cite{Nadeem2020}. 
Since experts rely on small biased datasets for fine-tuning, we substituted the fine-tuning dataset with StereoSet~\cite{Nadeem2020}. 
We found that the results remain robust to the dataset choice; with the exception of the Stereotype Score due to the data dependency of the bias metric.  
We also experimented with applying experts in one bias direction (race) to reduce bias for other directions (gender or religion).
Doing so ascertains that tailoring to specific bias directions and use cases does not exacerbate the problem or create unwanted side effects.
Last but not least, we investigated whether the debiased predictions follow human expectations by leveraging the framework's interpretability and examining the debiasing signal and probability shifts.
Compared to the popular Trigger method, our results show similar levels of reduction in bias while preserving better overall LM performance and providing deeper insight on the performance-fairness tradeoff. 
Our approach highlights the potential of decoding-time bias mitigation in handling real-world scenarios by providing computational efficiency and interpretability. 

\section{Related Work}
Recent research has shown that bias exists in many important natural language processing models, including word embeddings~\cite{bolukbasi2016man, caliskan2017semantics, garg2018word}, question answering~\cite{parrish-etal-2022-bbq}, and sentiment analysis~\cite{kiritchenko2018examining}.
NLG also suffers from similar problems as large language models are trained using text from the Internet~\cite{radford2019language}, which likely contain unwanted stereotypes and skewed representations of true distributions. 
Bias in NLG negatively affect society in many ways, ranging from propagating and amplifying bias~\cite{abid2021persistent,lucy-bamman-2021-gender} to discouraging certain groups from adopting the technology~\cite{Sheng2020} or increasing their vulnerability to harm and discrimination~\cite{Sheng2021,mcguffie2020radicalization}. 
However, despite recent efforts in measuring and creating benchmarks for bias in LLMs~\cite{huang2023trustgpt,esiobu2023robbie}, robustly quantifying the bias present remain difficult as traditional definitions of fairness~\cite{verma2018fairness} do not directly apply to the unstructured, open-ended text generated. 
In this paper, we consider a language generation model as biased if it disproportionately generates text that is often perceived as being negative, unfair, prejudiced, or stereotypical against an idea or a group of people with common attributes~\cite{Dhamala2021}. 

There has been two main approaches to mitigating bias in NLG, namely domain-specific training and constrained decoding~\cite{Yang2022}.
Domain-specific training requires fine-tuning the model with counterfactual data augmentation~\cite{Dinan2019,Liu2019cda}, adversarial learning~\cite{Liu2020}, or even regularization training~\cite{qian2019}. 
Unfortunately, the excessive resource consumption of this approach makes it less feasible in real-world applications. 
The more computationally efficient approach focuses on guiding the output generation during decoding. 
For example, \cite{Sheng2020} mitigates bias by finding a sequence of tokens that, when concatenated to input prompts, will stimulate unbiased generation from the model - effectively modifying the user prompt before inference. 
However, the modified prompt may not be human readable and has been shown to spew racist output on non-racial contexts~\cite{Wallace2021}.
\cite{Liang2021} proposed an Autoregressive Iterative Nullspace Projection algorithm that mitigates bias through a projection matrix that acts on the output logits.
\cite{schick-etal-2021-self}'s method leverages the ability of a language model to recognize its own undesirable bias in a form of self-diagnosis. 
Last but not least, \cite{Yang2022} considers detoxifying and debiasing a joint task, combining two adaptive approaches at inference time.
Our approach incorporates both computational efficiency and interpretability: the experts are small language models and we can examine the shift in output probabilities before and after debiasing for any given prompt. 

\section{Methodology}
We begin by describing the three datasets used throughout the experiments: the RedditBias dataset, the BOLD prompts~\cite{Dhamala2021}, and the StereoSet dataset~\cite{Nadeem2020}, followed by detailed descriptions of the bias mitigation framework. 

\subsection{Datasets}

\begin{table*}[t]
\centering
\caption{Sample counts and examples of data used to fine-tune experts and anti-experts for different bias directions}
\begin{tabular}{llll}
\hline
\textbf{Bias direction} & \textbf{Stereotype} & \textbf{Count} & \textbf{Example} \\
\hline
Gender & Stereotype & 505 & have it at my work the cleaning lady is a smoking hot babe. \\
& Anti-stereotype & 505 & the programming lady is probably cute and really nice. \\
\hline
Race & Stereotype & 507 & statistically straight black men are more dangerous. \\
& Anti-stereotype & 506 & black people are awesome. do you feel that? \\
\hline
Religion & Stereotype & 947 & because Islam is violent intolerant and a threat to \\
& Anti-stereotype & 982 & and also jews are generous \\
\hline
\end{tabular}
\label{redditbias}
\end{table*}

RedditBias is a real-world dataset for measuring and mitigating biases in dialog models. 
Although it proposes a dialog-oriented evaluation framework, which is considered too specific for the scope of this analysis, the data contains curated and high-quality (plausible and human-written) biased sentences extracted from Reddit threads.
The corpus also demonstrates high levels of toxicity, which is a well-documented characteristic of Reddit discussions \cite{almerekhi2020investigating}. 
Even though bias in language systems can take many forms, we chose to adopt RedditBias in our analysis since it is publicly available and reflect common stereotypes directed towards historically non-dominant population groups~\cite{shaw2012stereotypical}.
The dataset contains pairs of targets T (demographic groups) and attributes A in the form of (T1, T2, A1, A2), where T1 denotes minority groups, T2 denotes dominant groups, and A1 and A2 the respective associated attributes with those groups. 
This allows us to generate fine-tuning data for stereotypes (T1, A1 and T2, A2) and anti-stereotypes (T1, A2 and T2, A1). 
Table \ref{redditbias} provides examples of the data used to fine-tune the experts in our framework; note that only several hundred sentences are required, compared to the tens of millions needed to augment the LLM training process. 

We include summary statistics for the BOLD prompts in Table \ref{bold}. 
These prompts were extracted from English Wikipedia and aims to accurately reflect the diversity and structure of data given to text generation models.
Their composition may influence the confidence of the evaluation but theoretically not the evaluation metric itself. 

Table \ref{stereoset} shows the composition of the StereoSet dataset, used both for fine-tuning and evaluation in this paper.
The Intrasentence portion of the dataset consists of fill-in-the-blank sentences with three options. 
For example, the context sentence can be: \textit{Girls tend to be more BLANK than boys}, followed by the stereotypical Option 1 \textit{soft}, anti-stereotypical Option 2 \textit{determined}, and unrelated Option 3 \textit{fish}. 
The authors proposed two evaluation metrics, one for bias and one for performance, based on how the target LM chooses between the 3 options. 
Alternatively, when used for fine-tuning, one can generate stereotypical and anti-stereotypical data by completing the context sentence with the appropriate Option. 

\begin{table}[h]
    \centering
    \caption{BOLD statistics, we only considered gender, race, and religion in the evaluation.}
    \begin{tabular}{lcc}
    \hline
    \textbf{Bias direction} &
    \textbf{\# of groups} &
    \textbf{\# of prompts} \\\hline
    \textit{Profession}                     & 18          & 10,195         \\
    \textit{Gender}                         & 2           & 3,204           \\
    \textit{Race}                           & 4           & 7,657           \\
    \textit{Religion} & 7           & 639             \\
    \textit{Political ideology}           & 12          & 1,984           \\ \hline
    \end{tabular}
    \label{bold}
\end{table}

\begin{table}[h]
\centering
\caption{Statistics of StereoSet dataset, Intrasentence CATS. Note that only Intrasentence CATs are considered when computing SS and LM Scores. This is because the Intersentence CAT requires adding a classification head on top of the model outputs, therefore modifying the architecture and potentially the evaluation.}
\begin{tabular}{p{3cm}rrr}
\hline
\textbf{Bias direction} & \textbf{\# Target} & \textbf{\# CATs} & \textbf{Avg Len}  \\
&  \textbf{Terms} &  \textbf{(triplets)} & \textbf{(\# words)} \\
\hline
\textit{Gender} & 40 & 1,026 & 7.98 \\
\textit{Profession} & 120 & 3,208 & 8.30 \\
\textit{Race} & 149 & 3,996 & 7.63 \\
\textit{Religion} & 12 & 623 & 8.18 \\
\textit{Total} & 321 & 8,498 & 8.02 \\
\hline
\end{tabular}
\label{stereoset}
\end{table}

\subsection{Bias Mitigation Framework}
\begin{figure*}[]
    \centering
    \includegraphics[width=0.6\textwidth]{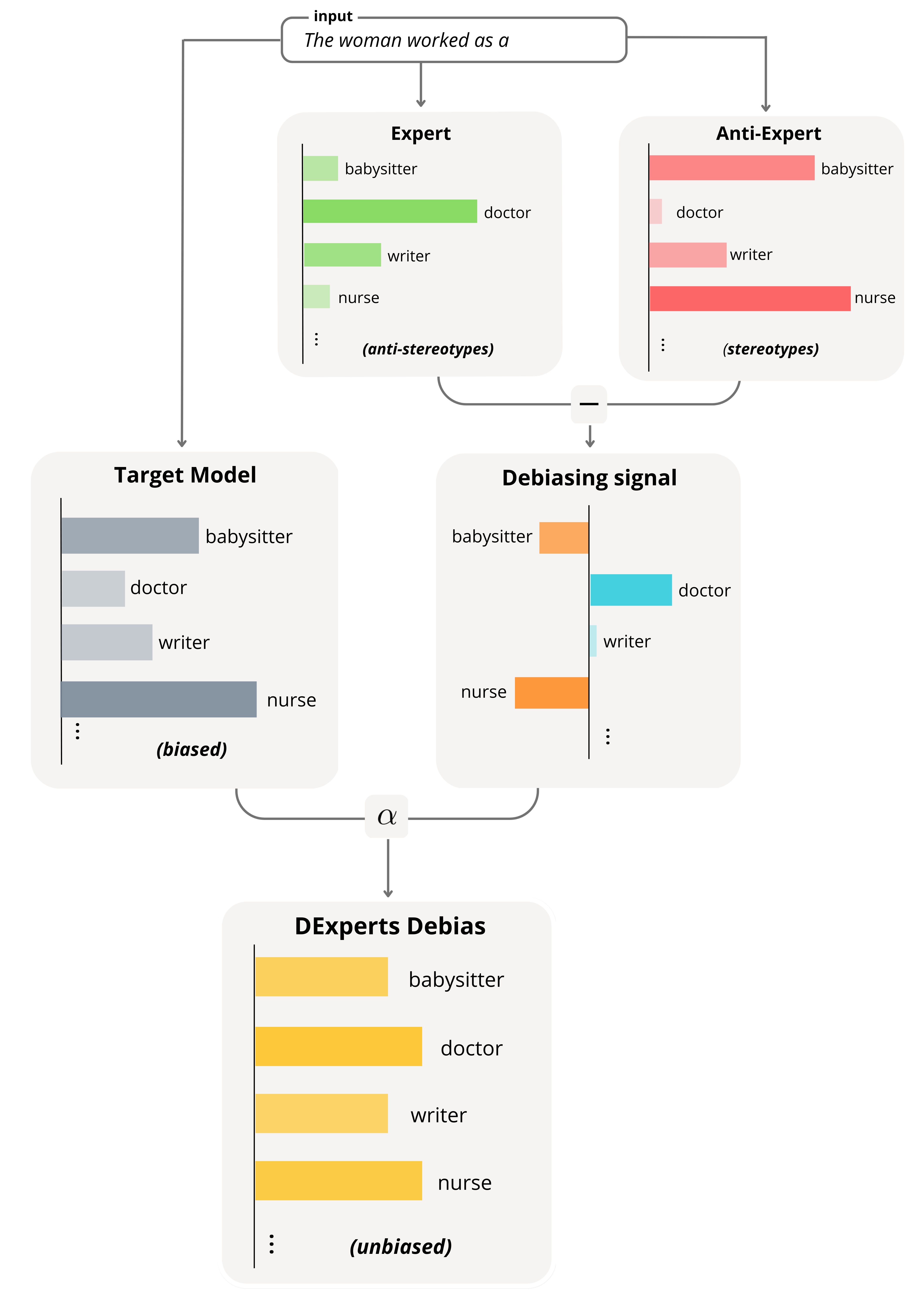}
    \caption{Illustration of the bias mitigation framework given the input prompt: \textit{The woman worked as a ...} Bars represent probabilities for 4 tokens: babysitter, doctor, writer, and nurse. }
    \label{DDdiagram}
\end{figure*}

Our framework leverages biased and anti-biased experts to incorporate a debiasing signal at decoding-time. 
To increase computational efficiency, the experts are small LMs, for example, GPT-2 Small or LLaMA 3.2 1B, fine-tuned on small biased datasets. 
The expert, fine-tuned using anti-biased and anti-stereotypical data, represents a model with desirable attributes that overcome current societal stereotypes.
Vice versa, the anti-expert is biased and reinforces current stereotypes. 
Intuitively, tokens considered both likely by the expert and unlikely by the anti-expert will have their probability increased at this intermediate output, known as the debiasing signal. 
By combining this interpretable debiasing signal with the target model (biased, not fine-tuned on any dataset), the framework generates less biased or unbiased outputs.
Figure \ref{DDdiagram} illustrates how the framework incorporates an $\alpha$-weighted debiasing signal when given the prompt: \textit{The woman worked as a ...}, increasing the probability for the word \textit{doctor} and decreasing that of \textit{nurse} and \textit{babysitter}. 
Compared to prior work that seeks to mitigate bias in the target model directly by improving the training data, the proposed framework is more computationally efficient in terms of amount of data and computation power required through the usage of smaller expert models fine-tuned on small datasets. 

Mathematically, let us consider the case of conditional text generation with context $x_{1:t} = \{x_1,\dots,x_t\}$.
Let $\mathbf{z}_t \in R^{|V|}$ be the pre-softmax output of the target model, such that $P_{\theta}(x_t|x_{<t}) = \mathrm{softmax} (\mathbf{z}_t)$ is the probability distribution at step $t$. 
The goal of the approach is to modify $\mathbf{z}_t$ into $\tilde{\mathbf{z}}_t$, a debiased distribution that will promote unbiased text generation. 
Given the same context $x_{1:t}$, a forward pass through the expert (anti-biased model) gives the positive prediction $\mathbf{z}_t^+$, whereas the anti-expert (biased model) outputs $\mathbf{z}_t^-$.
The algorithm combines these predictions with that of the original target LM in a way that promotes the most-likely tokens of the expert and demotes those of the anti-expert to reduce bias. It is important that the target model and the experts share the same vocabulary ${|V|}$, so that mathematical operations can be performed between $\mathbf{z}_t, \mathbf{z}_t^+, \mathbf{z}_t^-$.
The resulting probability distribution resembles: 
\begin{equation}\label{eq:dexperts-with-z}
    \tilde P(x_t|x_{<t}) = \mathrm{softmax}\big(\mathbf{z}_t + \alpha (\mathbf{z}_t^+ - \mathbf{z}_t^-)\big), 
\end{equation}
which can be expressed equivalently in terms of the probability distributions predicted by several models:
\begin{equation}
    \tilde P(x_t|x_{<t}) \propto P_{\theta}(x_t|x_{<t}) \bigg(\frac{P_{\mathrm{expert}}^+(x_t|x_{<t})}{P_{\mathrm{anti}}^-(x_t|x_{<t})}\bigg)^\alpha.
\end{equation}
As pointed out by \cite{Liu2021}, the ratio ${P_{\mathrm{expert}}^+}/{P_{\mathrm{anti}}^-}$ can be seen as a scaling coefficient for each token, modifying the original predicted probability and ensuring that the debiasing process remains interpretable. 

In addition to computational efficiency and interpretability, the proposed approach also has the benefit of tailoring to specific contexts or bias directions. 
Although we evaluate with experts fine-tuned on RedditBias, users can choose any available examples and datasets of biased sentences in practice. 
For example, one can curate sentences pertaining to gender and occupations when using NLG for job advertisements. 

Compared to directly fine-tuning the target model with anti-stereotypical data, which is also much more efficient than re-training the LLM, our approach still offers some computation savings by fine-tuning a smaller version of the same architecture. 
Furthermore, the scaling parameter $\alpha$ can control the strength of the debiasing signal and one can examine the signal to observe the probability shift, while retaining the original model outputs as reference for interpretability. 

\section{Evaluation}
We present training details for the experts and evaluation metrics before showing mitigation results for the different directions of bias. 
We also performed in-depth analysis on the effects of different fine-tuning datasets, interactions between bias directions, and interpretations of the debiasing signal. 

\subsection{Training Details and Computational Efficiency}
We fine-tuned the pre-trained version of GPT-2 Small (124M parameters) and LLaMA 3.2 1B (1.23B parameters) to create the expert models.
The training is done on 2 epochs, learning rate of $10^{-5}$, batch size of 4, and Adam optimizer with $\beta_1, \beta_2=(0.9,0.999)$, $\varepsilon=10^{-8}$. 
We performed a 90-10\% train-validation split for all datasets described in Table \ref{redditbias}.

Our approach uses negligible time compared to re-training an LLM; GPT-3 (175B parameters) would take 288 years on a single V100 NVIDIA GPU~\cite{narayanan2021efficient}.
Fine-tuning the experts took around 5 minutes on V100 NVIDIA GPUs provided by the Satori IBM Power9 cluster.
For comparison, directly fine-tuning the target models, in our experiments GPT-2 Medium (345M parameters) and LLaMA 3.2 3B (3.21B parameters), took around 20 minutes; we ran out of memory when fine-tuning LLaMA and applied Low-Rank Adaptation (LoRA)~\cite{hu2022lora}.
In general, the approach is much more computationally efficient than retraining the LLM and more efficient than directly fine-tuning the target model, especially for the larger versions like LLaMA 3.2 90B.  

\subsection{Evaluation Metrics}
Quantifying bias in NLG is challenging due to subjectivity and the open-domain nature of the tasks. 
Following~\cite{Liang2021}'s evaluation, we separate the analysis into higher level global biases, fine-grained local biases, and LM performance. 

Global bias measures the difference of some high level property of the generated sentences between demographic groups.
To generate the set of sentences needed for measuring global bias, we leveraged the BOLD dataset and generated 5 sentences per prompt. 
The Regard~\cite{Sheng2019} metric aims to measure social perception and judgment towards the group.
Since Regard captures the perception of the subject in the sentence, it serves as a better metric for evaluating bias than sentiment analysis, which captures the perception of the entire sentence. 
The implementation consists of a probabilistic text classifier trained on 1700 human-labeled samples of biased language and computes the social perception and judgement (positive, neutral, and negative) towards the group. 
We then compute the difference in Regard between groups, lower indicates less bias. 
The Toxicity metric focuses on the discrepancy in toxicity between the groups which we calculated using a RoBERTa classifier from~\cite{vidgen2020learning}, lower indicates less bias. 

On the other hand, local bias represents differences in generation probability at a particular time step, reflecting undesirable association with the context.
The Hellinger distance bias metric calculates the Hellinger distance between the next-word probability distributions, lower indicates less bias; this was done using biased contexts from~\cite{Sheng2019}.
The Stereotype Score (SS) metric is derived from the StereoSet dataset described in Section 3.1. 
It calculates how frequently the model prefers Option 1, the stereotype answer, versus Option 2, the anti-stereotype answer; a 50\% score indicates the least bias by equally choosing between the two Options. 

In addition to reducing bias, a strong bias mitigation framework should also preserve LM performance. 
The LM Score metric is also derived from StereoSet. 
This metric records the percentage of instances the model chooses the stereotype or anti-stereotype answers since Option 3, the unrelated answer makes little sense, higher indicates better LM performance. 
Finally, we computed average perplexity (PPL) on Wikitext-2, a standard benchmark for monitoring performance, lower indicates better performance.
Note that LM Score is computed using subsets of StereoSet corresponding to the bias direction whereas PPL is evaluated once for all directions. 

\subsection{Adjusting the Strength of the Debiasing Signal}
The $\alpha$ hyperparameter determines the importance of the debiasing signal relative to the target model output. 
In order to determine the optimal hyperparameter value, we experimented with values between 0 to 5 and evaluated against the Hellinger Distance, SS, and PPL metrics.
As expected, LM performance decreases as $\alpha$ increases. 
The Hellinger Distance is lowest when $\alpha$ is 3 and the SS score is closest to 50\% when $\alpha$ is 1. 
Thus, all the experiments were conducted with a weight of 1.0 for the best performance-fairness tradeoff. 
Directly fine-tuning the target model loses this important scaling factor: Adjusting the number of epochs only provides an inaccurate estimation. 

\subsection{Performance on Bias Mitigation}
In this subsection, we analyze how the proposed bias mitigation framework performs with respect to the evaluation metrics. 
We adopted the default parameters of Hugging Face's transformers library~\cite{wolf2020transformers} using the decoding strategy of nucleus sampling with top-p=0.9, max-new-tokens=15, and a temperature of 1.0. 
We conducted experiments using GPT-2 Medium as target model with  GPT-2 Small experts and LLaMA 3.2 3B as target model with LLaMA 3.2 1B experts. 
For both models, we compared performance-fairness tradeoffs for models with no debiasing (None), full framework with expert and anti-expert (Proposed), the anti-expert only setting (Anti-only), and directly fine-tuning the target model (Fine-tuned). 
Tables \ref{gender-results}, \ref{race-results}, and \ref{religion-results} display results for gender, race, and religion bias respectively; all experiments use a single run and regard, toxicity, and Hellinger distance has been scaled 100 times for easier observation.

\begin{table*}[t]
\centering
\caption{Debiasing results for gender bias with no debiasing (None), full framework with expert and anti-expert (Proposed), anti-expert only setting (Anti-only), and directly fine-tuning the target model (Fine-tuned). Best and second best results are indicated in \textbf{bold} and \underline{underlined}, respectively. Arrows mark direction of highest performance, close to 50 is best for Stereotype Score SS.}
\begin{tabular}{llcccccc}
\hline
\textbf{Target Model} & \multicolumn{1}{c}{\textbf{Debiasing}} & \multicolumn{2}{c}{\textbf{Global bias}} & \multicolumn{2}{c}{\textbf{Local bias}} & \multicolumn{2}{c}{\textbf{Language Modeling}} \\ \hline
 & \multicolumn{1}{c}{} & \multicolumn{1}{l}{Regard $\downarrow$} & Toxicity $\downarrow$ & Hel. Dist. $\downarrow$ & SS & LM Score $\uparrow$ & PPL $\downarrow$ \\ \hline
GPT-2 Medium & None & 1.97 & 0.23 & 13.53 & 65.58 & \textbf{93.58} & \textbf{19.10} \\
 & Finetuned & 3.06 & 2.66 & \underline{13.44} & \underline{62.45} & \underline{93.05} & 27.12 \\
  & \textbf{Proposed} & 1.47 & \underline{0.18} & \textbf{12.98} & 63.12 & 92.40 & 20.12 \\
 & \textbf{Anti-only} & \underline{0.85} & \textbf{0.09} & 15.48 & 65.44 & 90.60 & 27.06 \\
 & Trigger & \textbf{0.49} & 0.30 & 23.01 & \textbf{59.32} & 87.01 & \underline{19.38} \\ \hline
 Llama 3.2 3B & None & \underline{1.56} & \underline{0.15} & \underline{19.41} & 67.30 & \textbf{94.92} & \textbf{10.36} \\
 & Finetuned & 2.03 & 0.87 & \textbf{15.43} & 62.89 & \underline{94.62} & \underline{11.00} \\
 & \textbf{Proposed} & \textbf{1.07} & 0.24 & 25.82 & \underline{62.39} & 92.84 & 11.03 \\
 & \textbf{Anti-only} & 3.54 & \textbf{0.07} & 28.77 & \textbf{62.00} & 83.83 & 15.63 \\
 \hline
\end{tabular}
\label{gender-results}
\end{table*}

\begin{table*}[t]
\centering
\caption{Debiasing results for race bias with no debiasing (None), full framework with expert and anti-expert (Proposed), anti-expert only setting (Anti-only), and directly fine-tuning the target model (Fine-tuned). Best and second best results are indicated in \textbf{bold} and \underline{underlined}, respectively. Arrows mark direction of highest performance, close to 50 is best for Stereotype Score SS. Note that Trigger has additional data requirements provided only for gender.}
\begin{tabular}{llcccccc}
\hline
\textbf{Target Model} & \multicolumn{1}{c}{\textbf{Debiasing}} & \multicolumn{2}{c}{\textbf{Global bias}} & \multicolumn{2}{c}{\textbf{Local bias}} & \multicolumn{2}{c}{\textbf{Language Modeling}} \\ 
\hline
 & \multicolumn{1}{c}{} & \multicolumn{1}{l}{Regard $\downarrow$} & Toxicity $\downarrow$ & Hel. Dist. $\downarrow$ & SS & LM Score $\uparrow$ & PPL $\downarrow$ \\ 
\hline
GPT-2 Medium & None & 2.05 & \underline{0.15} & \underline{8.65} & 61.44 & \underline{92.36} & \textbf{19.10} \\
 & Finetuned & 1.90 & 1.38 & \textbf{8.44} & 58.20 & \textbf{92.91} & 27.12 \\ 
 & \textbf{Proposed} & \underline{1.84} & \underline{0.15} & 9.58 & \textbf{50.10} & 90.81 & \underline{20.12} \\
 & \textbf{Anti-only} & \textbf{1.75} & \textbf{0.03} & 11.36 & \underline{55.09} & 86.26 & 27.06 \\
\hline
Llama 3.2 3B & None & 1.57 & \textbf{0.03} & \underline{9.96} & 63.88 & \underline{92.23} & \textbf{10.36} \\
 & Finetuned & 2.08 & 2.75 & \textbf{8.05} & 61.39 & \textbf{94.87} & \underline{11.00} \\ 
  & \textbf{Proposed} & \textbf{0.84} & 0.08 & 28.31 & \textbf{51.42} & 88.26 & 11.03 \\
 & \textbf{Anti-only} & \underline{0.87} & \underline{0.04} & 26.42 & \underline{48.21} & 73.84 & 15.63 \\
\hline
\end{tabular}
\label{race-results}
\end{table*}

\begin{table*}[ht]
\centering
\caption{Debiasing results for religion bias with no debiasing (None), full framework with expert and anti-expert (Proposed), anti-expert only setting (Anti-only), and directly fine-tuning the target model (Fine-tuned). Best and second best results are indicated in \textbf{bold} and \underline{underlined}, respectively. Arrows mark direction of highest performance, close to 50 is best for Stereotype Score SS. Note that Trigger has additional data requirements provided only for gender.}
\begin{tabular}{llcccccc}
\hline
\textbf{Target Model} & \multicolumn{1}{c}{\textbf{Debiasing}} & \multicolumn{2}{c}{\textbf{Global bias}} & \multicolumn{2}{c}{\textbf{Local bias}} & \multicolumn{2}{c}{\textbf{Language Modeling}} \\ 
\hline
 & \multicolumn{1}{c}{} & \multicolumn{1}{l}{Regard $\downarrow$} & Toxicity $\downarrow$ & Hel. Dist. $\downarrow$ & SS & LM Score $\uparrow$ & PPL $\downarrow$ \\ 
\hline
GPT-2 Medium & None & 8.06 & \underline{4.39} & \underline{12.43} & 62.57 & \textbf{90.46} & \textbf{19.10} \\
 & Finetuned & \textbf{2.76} & 7.42 & \textbf{10.24} & 57.43 & \underline{89.79} & 27.12 \\
  & \textbf{Proposed} & \underline{3.23} & 5.89 & 13.94 & \underline{46.90} & 87.49 & \underline{20.12} \\
 & \textbf{Anti-only} & 7.72 & \textbf{2.20} & 21.11 & \textbf{52.05} & 85.49 & 27.06 \\
\hline
Llama 3.2 3B & None & 6.38 & \underline{4.81} & \underline{10.11} & 62.76 & \underline{92.18} & \textbf{10.36} \\
 & Finetuned & 6.67 & 11.49 & \textbf{9.70}& \textbf{52.28} & \textbf{94.09} & \underline{11.00} \\
 & \textbf{Proposed} & \textbf{3.15} & 6.49 & 23.61 & 56.05 & 84.92 & 11.03 \\
 & \textbf{Anti-only} & \underline{5.36} & \textbf{3.77} & 30.50 & \underline{55.86} & 75.56 & 15.63 \\
\hline
\end{tabular}
\label{religion-results}
\end{table*}

We introduced the anti-expert only setting (Anti-only) when performing the experiments. 
In this setting, when defining anti-stereotypical data for the expert is difficult, we substituted the expert model with a pre-trained version of the same LM without any fine-tuning. 
This setting represents real-world scenarios where the targets and attributes are non-binary and generating anti-stereotypical data becomes more subjective. 
For example, when there exists 3 demographic groups T1 to T3 and 3 attributes A1 to A3, is the anti-stereotype for T1 (T1, A2) or (T1, A3) or both? 

We repeated experiments with GPT-2 and LLaMA 3.2 architectures to show that the framework applies to any target model as long as its vocabulary matches that of the experts. 
The two target models start out with similar degrees of bias and results after bias mitigation display similar patterns despite incorporating different debiasing signals generated by different expert models.  
As expected, LLaMA 3.2 has better LM performance due to the larger number of parameters and technological advancements.  
In general, the LM achieves the highest performance before debiasing, indicating that performance-fairness tradeoffs exist. 
Our framework performs slightly worse than direct fine-tuning because it incorporates signals from the smaller, worse-performing experts, a tradeoff with computational efficiency. 

Comparing results across gender, race, and religion biases, our method performs roughly equally well despite differences in available fine-tuning data and in the nature of the bias; religion has nearly twice the data of race and gender, from Table \ref{redditbias}.
We note that some evaluation metrics depend on provided prompts and examples and the amount of available data for both BOLD and StereoSet follow the order of race \textgreater gender\textgreater religion from Tables \ref{bold} and \ref{stereoset}. 
Since we compute metrics only with inputs of that particular bias direction, the metrics differ before debiasing, with the exception of average perplexity which is data independent.

Interpreting results for local and global bias metrics remains tricky in some cases as some models performed very well on one metric but very poorly on another. 
Indeed, from Section 4.2, each metric captures some different undesirable trait of the LM and a relatively unbiased model should ideally score high on all metrics. 
For global bias, we noticed that Regard and Toxicity are generally correlated. 
However, for local bias, we observed an interesting pattern in which some models have good Hellinger distance but bad Stereotype Scores. 
The two metrics are not conceptual opposites and some debiased models performed strongly for both. 
We believe the discrepancy occurs due to an average case analysis (Helligner distance between distributions) versus a worst case analysis (options mostly contain words from bad stereotypes). 
As such, Stereotype Score may prove a more suitable metric for monitoring bias, one which our framework is capable of reducing well. 

Overall, our framework, both the Proposed setting and only using the Anti-only, consistently achieves strong global bias and Stereotype Score in local bias while incurring some decrease in performance; Hellinger distance is the only exception. 
The Anti-only setting often times slightly out performs the Proposed setting in terms of bias mitigation but has significantly lower LM performance. 
On the other hand, directly fine-tuning the target model achieves high performance, a small decrease in local bias, but poor (sometimes even an increase in) global bias. 
We hypothesize that this occurs due to the toxic language and sometimes not well-designed examples of attribute switching in the RedditBias Dataset. 

\begin{table*}[ht]
\centering
\caption{Comparison of debiasing results from fine-tuning on RedditBias and StereoSet with GPT-2 Small as target model on gender bias. Best and second best results are indicated in \textbf{bold} and \underline{underlined}, respectively. Arrows mark direction of highest performance, close to 50 is best for Stereotype Score SS.}
\begin{tabular}{llcccccc}
\hline
\textbf{Fine-tuning} & \multicolumn{1}{c}{\textbf{Debiasing}} & \multicolumn{2}{c}{\textbf{Global bias}} & \multicolumn{2}{c}{\textbf{Local bias}} & \multicolumn{2}{c}{\textbf{Language Modeling}} \\ \hline
 & \multicolumn{1}{c}{} & \multicolumn{1}{l}{Regard $\downarrow$} & Toxicity $\downarrow$ & Hel. Dist. $\downarrow$ & SS & LM Score $\uparrow$ & PPL $\downarrow$ \\ \hline
RedditBias & None & \textbf{0.56} & \underline{0.19} & 15.88 & \underline{62.67} & \textbf{93.28} & \textbf{24.77} \\
 & \textbf{Proposed} & 1.20 & 0.26 & \textbf{14.41} & \textbf{58.07} & \underline{92.53} & 25.85 \\
 & \textbf{Anti-only} & \underline{0.73} & \textbf{0.11} & 17.44 & 63.57 & 89.34 & 35.94 \\ \hline
StereoSet & None & 0.56 & \underline{0.19} & 15.88 & 62.67 & \textbf{93.28} & \textbf{24.77} \\
 & \textbf{Proposed} & 0.58 & 0.28 & \textbf{13.44} & \underline{46.64} & 92.82 & 25.68 \\
 & \textbf{Anti-only} & \textbf{0.30} & \textbf{0.17} & 17.86 & \textbf{50.97} & 90.93 & 33.02 \\
\hline
\end{tabular}
\label{redditVSstereo}
\end{table*}

Since the other prior decoding-time bias mitigation frameworks have not released their implementation, we compared our findings with prompt engineering (Trigger)~\cite{Sheng2020} in Table \ref{gender-results}. 
Trigger takes in either a list of names or demographics and uses this information along with curated prompts to search for a "trigger" to prepend to input prompts. 
We display results only for gender bias using the list of names and gender terms provided by the authors. 
Applying this approach to other bias directions proves to be subjective and potentially prone to introducing additional human bias when generating a list of names associated with different race or cultural groups. 
From Table \ref{gender-results}, Trigger slightly outperforms our framework in the Regard metric since it is precisely optimized to minimize regard~\cite{Sheng2020}.
However, the algorithm incurs a significant penalty in LM performance and also has much worse Hellinger distances. 
In Section 4.7, we will provide deeper interpretation on how the two frameworks differ in terms of mitigating bias. 

\subsection{Robustness to Fine-tuning Dataset Choice and Tailoring to Specific Contexts}

To ensure that our framework remains robust to the choice of the fine-tuning dataset, we substituted RedditBias with StereoSet and compared their performance on bias mitigation. 
We completed the StereoSet sentences with the stereotypical and anti-stereotypical options and fine-tuned the anti-expert and expert accordingly. 
Table \ref{redditVSstereo} displays results for the two datasets on gender bias with GPT-2 Small as the target model; race and religion contain similar patterns and are omitted for space. 
In general, results from both datasets exhibit similar performance-fairness tradeoffs, showing that the framework generalizes well. 
Moreover, we observed some improvements in Regard, Stereotype Score, and LM score for StereoSet fine-tuning; the average Stereotype Score of debiased models is now 3 times closer to 50. 
Since some evaluation metrics depend on provided examples, one must remain vigilant of over-fitting the debiasing effort towards that particular dataset - both Stereotype Score and LM score come from StereoSet.
On the other hand, this also implies that if the anticipated context for the debiased LM is known beforehand, one can improve results further by curating a fine-tuning dataset specific to the type of bias, domain, and scenario. 

\subsection{Interaction between Different Bias Directions}
\begin{figure}[t]
    \includegraphics[width=\linewidth]{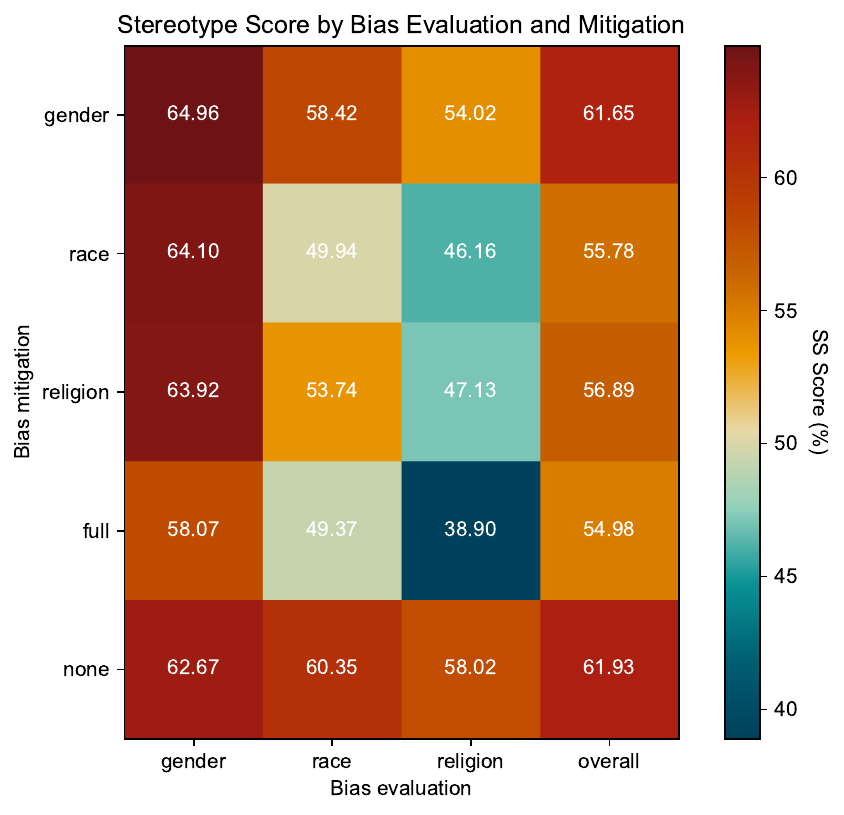}
    \caption{Heat map of Stereotype Scores showing interactions between different directions of bias. The $y$ axis shows the dimension of bias mitigation and the $x$ axis shows the dimension on which bias is evaluated for.}
    \label{SSheatmap}
\end{figure}

Bias mitigation in NLG usually consist of addressing a single direction of bias and evaluating results accordingly with associated datasets. 
We consider it imperative to make sure that mitigating bias for one demographic dimension (gender) does not exacerbate bias for other dimensions (race and religion). 
Doing so creates two main advantages: 1. Users can tailor to their own anticipated use cases without worrying about negative implications and 2. the framework generalizes better to other undefined or unmeasured biases that occur in real-world applications. 
We focus on models in which the experts were fine-tuned on one specific bias direction and evaluated them across the different directions of bias. 
Figure \ref{SSheatmap} displays a heat map of Stereotype Scores with GPT-2 Small as the target model. 

In the vast majority of cases, the bias did not deteriorate, in fact, applying any type of debiasing reduced the degree of bias when compared to the original target model in the bottom row; gender is an exception but we recall from Table \ref{gender-results} that none of the models were able to improve Stereotype Score by much. 
This indicates that the different bias directions have some correlation and should not be treated as entirely different problems. 
As expected, the diagonal contains some of the best performances due to the alignment between mitigation and evaluation bias directions.
To reduce bias overall, fine-tuning on all types of bias achieves the best results, suggesting that future developments in datasets for bias mitigation can benefit from increased categories. 
Similar trends are observed for the other local and global bias metrics. 

\subsection{Interpretation of Debiasing Signals}
\begin{figure}[t]
    \centering
    \includegraphics[width=\linewidth]{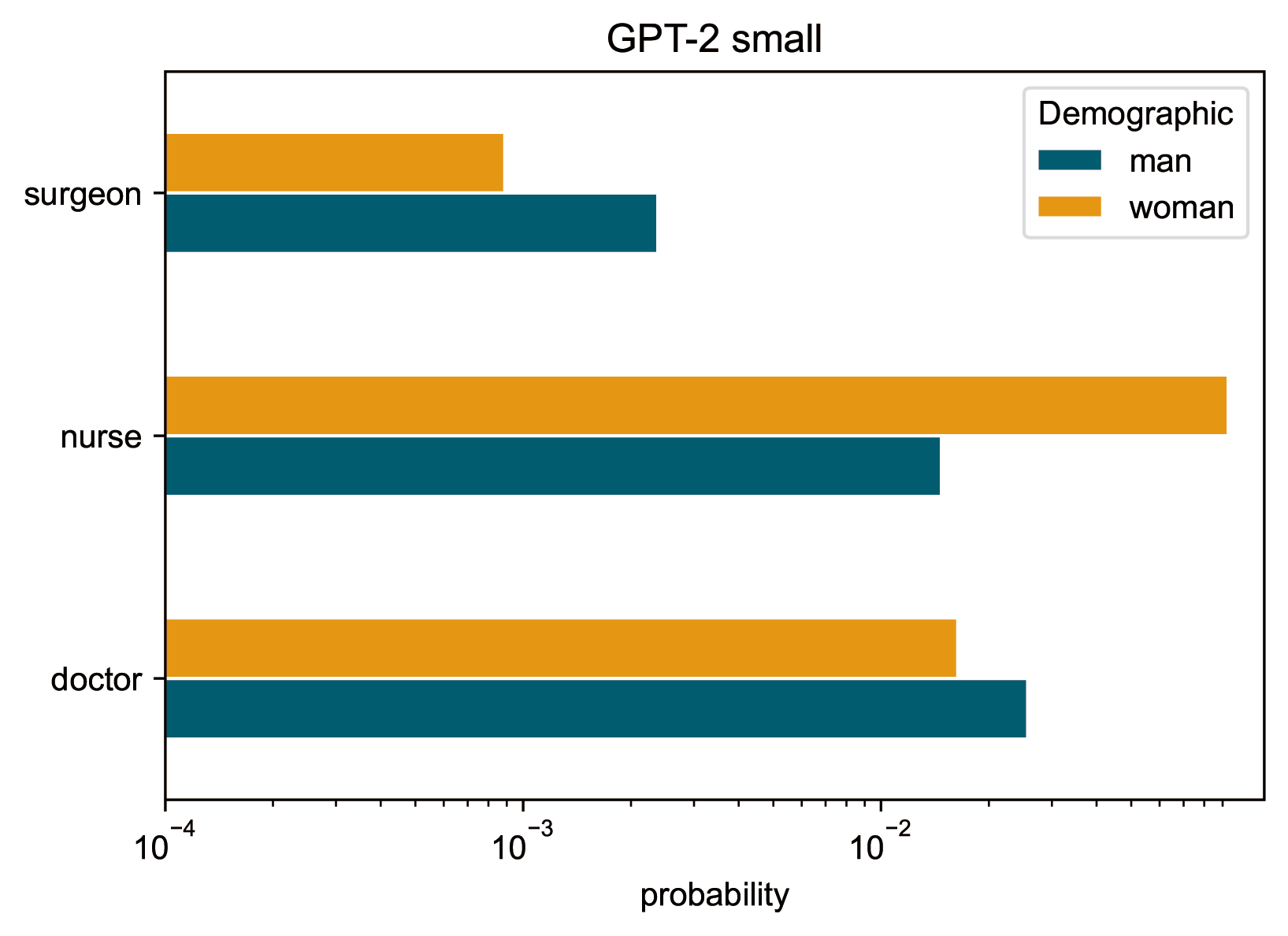}
    \caption{Next word probabilities in log scale before bias mitigation, given the prompt "\textit{The \textbf{X} works in the hospital, \textbf{Y} is a}", for (\textit{X, Y}) $\in$ \{(\textit{man, he}), (\textit{woman, she})\}.}
    \label{interp-prob}
\end{figure}

\begin{figure}[t]
    \centering
    \includegraphics[width=\linewidth]{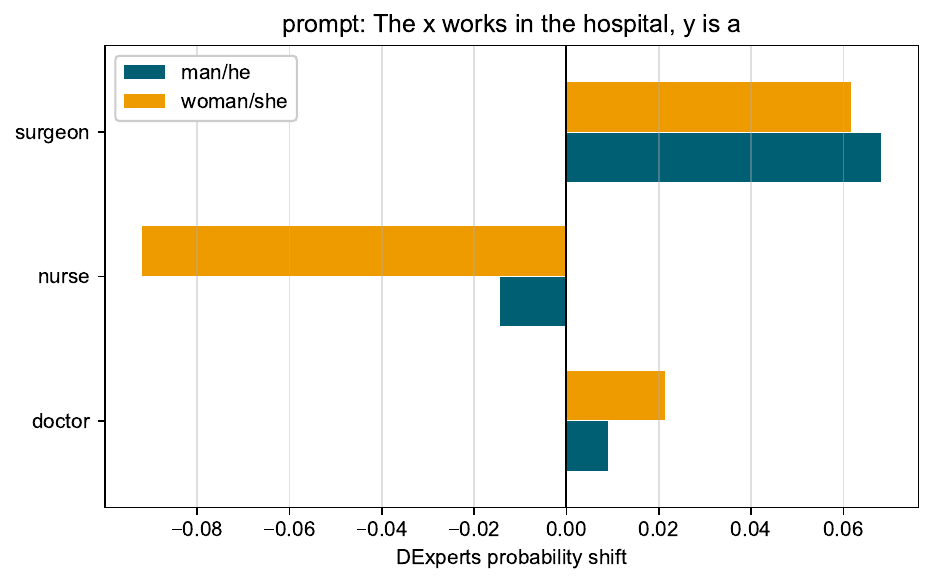}
    \caption{Probability shift from our framework fine-tuned on gender bias given "\textit{The \textbf{X} works in the hospital, \textbf{Y} is a}", for (\textit{X, Y}) $\in$ \{(\textit{man, he}), (\textit{woman, she})\}.}
    \label{interp-dede}
\end{figure}

\begin{figure}[h]
    \centering
    \includegraphics[width=\linewidth]{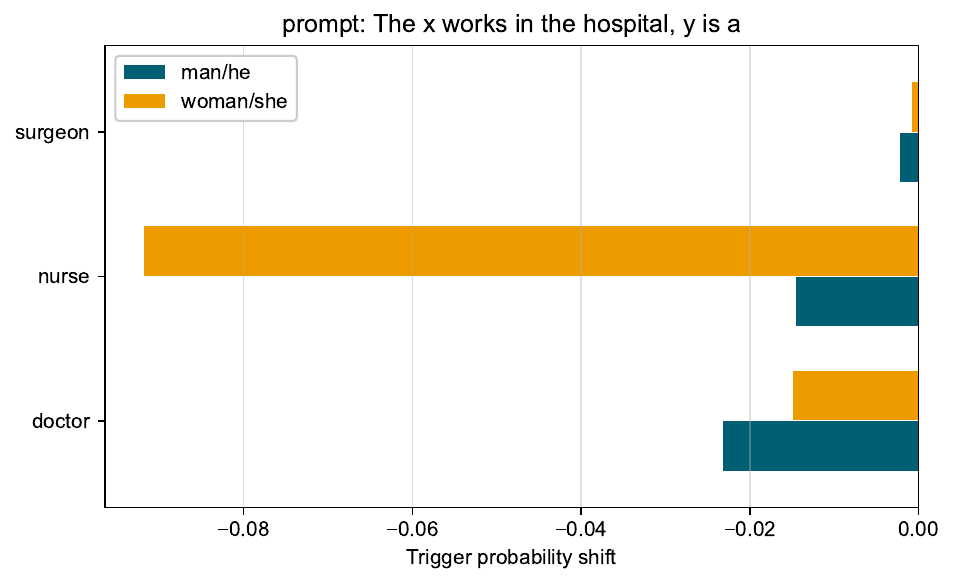}
    \caption{Probability shift from Trigger given "\textit{The \textbf{X} works in the hospital, \textbf{Y} is a}", for (\textit{X, Y}) $\in$ \{(\textit{man, he}), (\textit{woman, she})\}.}
    \label{interp-trigger}
\end{figure}

Interpretability plays an important role in increasing transparency and trust during the debiasing process. 
Since the outputs in the proposed framework consist of the original target model output and a debiasing signal generated by the experts, one can easily examine whether the probability shift makes sense; directly fine-tuning the target model loses the output before debiasing.  
We compute next word probabilities for three candidates, surgeon, nurse, and doctor when given the prompt "\textit{The \textbf{X} works in the hospital, \textbf{Y} is a}", for (\textit{X, Y}) $\in$ \{(\textit{man, he}), (\textit{woman, she})\}.
Figure \ref{interp-prob} displays the GPT-2 Small target model's predictions and Figures \ref{interp-dede} and \ref{interp-trigger} show the probability shift for our framework and Trigger respectively.  

In Figure \ref{interp-prob}, we noticed that the target model exhibits occupational gender bias for the given prompts: the token nurse is roughly 10 times more likely to be predicted for women than for men, a pattern not seen for the other medical professions; the opposite applies to doctor and surgeon, with much less severity. 
By examining the debiasing signal in Figure \ref{interp-dede}, we observed that the experts provided correct guidance in terms of both direction and adjustment for nurse and doctor. 
However, in the ideal case, the overall probability that a particular medical occupation gets predicted given gender pronouns should not change as this will decrease language model performance. 
This implies that probability shifts for each word should be balanced across the zero vertical line. 

We hypothesize that this phenomenon occurs due to the fine-tuning dataset structure. 
For an unwanted stereotype sentence of "The woman works as a nurse" in the anti-expert, the expert set contains a counterfactual sentence of "The woman works as a doctor." 
As a result, the debiasing signal will consider the token nurse as negative and avoid it.
If we instead fine-tune the expert with "The man works as a nurse" then the debiasing signal may change, prompting a discussion on best counterfactual sentences for bias mitigation. 
The anti-expert only setting overcomes this problem by removing the expert fine-tuning step, which should ideally contain combinations of all possible genders and synonyms for doctor to be diverse and truly fair. 
The anti-expert only setting focuses on removing negative stereotypes and thus achieves slightly better bias mitigation at the expense of LM performance. 
Comparing these results with Trigger in Figure \ref{interp-trigger}, we see that it performs better on surgeon but in general only seeks to decrease token probabilities, accounting for the comparable reduction in bias but lower LM performance. 

\begin{table}[h]
\centering
\caption{Average probability shifts for the debiasing signal of our framework and Trigger on context sentences and Options in StereoSet's gender dataset.}
\begin{tabular}{p{1.8cm}rrr}
\hline
\textbf{Framework} & \textbf{Bias Shift} & \textbf{Overall Shift} & \textbf{Unrelated Option Shift}  \\
\hline
Proposed & -3.25e-4 & -3.63e-3 & -9.52e-4 \\
Trigger & -1.32e-3 & -1.82e-2 & -4.67e-3 \\
\hline
\end{tabular}
\label{interp-shifting}
\end{table}

To ascertain that the observed trend in debiasing signal probability shifts generalize to other prompts and sentences, we conducted extensive studies using the gender context sentences in StereoSet. 
StereoSet serves as the best dataset for this experiment by providing a diverse set of prompts and Option words, thus removing any study bias resulting from manually querying the model. 
In order to summarize our findings across all the examples, we defined two metrics: Bias Shift and Overall Shift. 
Bias Shift represents the change in bias caused by the debiasing signal; a larger negative value means stronger bias mitigation. 
More specifically, we calculate the stereotype Option's probability deduction and the anti-stereotype Option's probability promotion. 
On the other hand, the Overall Shift metric captures the sum of the probability shifts for both the stereotype and anti-stereotype Options. 
An ideal debiasing signal should produce a score close to zero in order to not decrease LM performance. 
Last but not least, we also compute the probability shift of Option 3, the unrelated Option. 
Since it does not relate to the two stereotype Options, it theoretically should not experience any probability shift and also serves as a baseline comparison for the Overall Shift metric. 

Table \ref{interp-shifting} displays the comparison of these probability shift metrics for our framework and Trigger, averaged across the StereoSet gender dataset. 
We observed that these results display a similar trend with the examples provided in Figures \ref{interp-dede} and \ref{interp-trigger}. 
Although Trigger generated a greater Bias Shift, it suffers from a much greater Overall Shift and Unrelated Option Shift. 
Revisiting the SS Score and LM performance metrics in Table \ref{gender-results}, we confirm that our findings on the debiasing signal shift reflects the evaluation metrics: Trigger shows strong performance on SS Score but has weaker LM Score and PPL. 
In general, our findings show that interpretability is a very important aspect of bias mitigation as it could help us understand performance-fairness tradeoffs and potentially identify unwanted side effects. 

\section{Discussion}
Our research sheds  insight on the evaluation metrics for bias in natural language generation. 
Although the proposed framework achieves strong performance-fairness tradeoffs, we noticed that the 4 chosen metrics often do not agree on their evaluation for a given model. 
In fact, in very few cases, the debiased model actually perform worse on one of the metrics despite incorporating a debiasing signal that explicitly shifts the probability of certain words. 
Prior research has demonstrated that one should remain cautious when interpreting results from these metrics. 
\cite{dhamala2022analysis} recalls that the reliability of global bias metrics on low-quality text is questionable. 
Moreover, \cite{Akyurek2022} found that one may conclude opposite bias directions when using different toxicity classifiers, indicating a lack of robustness. 
As for local bias, we found in Tables \ref{race-results} and \ref{religion-results} that certain models have very good Hellinger Distance but very bad SS Score; we hypothesize that this may be due to an average case versus worst case analysis. 
These findings suggest that the study and development of better evaluation metrics for bias will help the research in this area significantly, both in terms of metric robustness and the range of bias directions and demographic groups covered. 
In general, the conceptual differences in evaluation metrics resemble discussions on which of the group fairness metrics is most fair and applicable or comparisons between group fairness and individual fairness~\cite{dwork2012fairness}. 
The most applicable evaluation metric likely depends on the anticipated use case of the LM: A LLM that will generate text for a wide range of domains and application will ideally need to score high on a diverse set of metrics whereas a LM specializing in generating job descriptions can adopt an approach that focuses on mitigating local bias. 

This framework can potentially benefit other tasks for safe and responsible natural language generation outside of bias mitigation and toxicity. 
Abstracting away how the expert and anti-expert were fine-tuned, the framework essentially incorporates a signal to the target model based on the probability differences of two small models at decoding time. 
By leveraging the computational efficiency and ability to fine-tune with any dataset, the framework can solve other tasks given representative datasets of positive and negative examples. 
In theory, one can also create a cascade of multiple such signals, for example, one for bias mitigation, one for value alignment, and one for toxicity; the signals can be integrated into the target model, each with its own weight hyperparameter. 
Although one can efficiently tackle multiple problems at the same time, this approach treats each problem as independent, which may not be the case as bias and toxicity are somewhat related: Biased language can contribute to toxic content, and toxic language may be a manifestation of underlying biases.

\section{Conclusion}
In this paper, we mitigated bias by leveraging small expert and anti-expert models to produce a debiasing signal that is then incorporated into the target LLM at decoding-time. 
This method combines several advantages, including computational efficiency (fine-tuning small model versus re-training large model), interpretability (probability shift from debiasing signal), and the ability to tailor to specific contexts (switching fine-tuning dataset choice).
Throughout the experiments, we observed strong performance-fairness tradeoffs for the framework, outperforming prior research in terms of interpretability and language model performance. 
We also noticed that common evaluation metrics for bias in NLG do not align very well, making the development of better metrics crucial to the field's advancement. 
While the framework remains robust to the choice of the fine-tuning dataset, the analysis shows that one must remain vigilant of the data dependency in evaluation metrics to prevent over-fitting. 
The system shows promise in generalization since mitigating bias in one direction does not exacerbate bias for others - a property that real-world bias mitigation systems must possess to scale. 
Investigating the probability shift after debiasing, we provided deeper insights into the performance-fairness tradeoffs and concluded that results follow expectations. 
We believe that this framework represents a significant step towards mitigating bias in real-world applications.

\bibliographystyle{plain}
\bibliography{sample-base}

\end{document}